\pdfoutput=1

\documentclass[11pt]{article}

\usepackage[]{acl}

\usepackage{times}
\usepackage{latexsym}

\usepackage[T1]{fontenc}

\usepackage[utf8]{inputenc}

\usepackage{microtype}

\usepackage{inconsolata}

\usepackage{amsmath,amsfonts,bm}

\def\eqref#1{equation~\ref{#1}}

\def\1{\bm{1}}

\DeclareMathAlphabet{\mathsfit}{\encodingdefault}{\sfdefault}{m}{sl}
\SetMathAlphabet{\mathsfit}{bold}{\encodingdefault}{\sfdefault}{bx}{n}

\usepackage{hyperref}
\usepackage{url}

\usepackage{booktabs}
\usepackage{pifont}

\usepackage{multirow}
\usepackage{multicol}

\usepackage{colortbl}
\usepackage[linewidth=0.8pt]{mdframed}

\usepackage[linesnumbered, ruled,vlined]{algorithm2e}
\usepackage{algorithmic}
\usepackage{graphicx}
\usepackage{wrapfig}
\usepackage{soul}
\usepackage{tcolorbox}

\newcommand{\hs}[1]{\textcolor{magenta}{{#1}}}
\newcommand{\zts}[1]{\textcolor{red}{Tianshu: #1}}
\newcommand{\lyf}[1]{\textcolor{blue}{[Yifei: #1]}}

\newcommand{\nop}[1]{}

\title{TableLlama: Towards Open Large Generalist Models for Tables}

\author{Tianshu Zhang\quad Xiang Yue\quad Yifei Li\quad Huan Sun\\
The Ohio State University\\
\texttt{\{zhang.11535,yue.149,li.14042,sun.397\}@osu.edu}
\\
}

\begin{document}
\maketitle

\begin{abstract}
Semi-structured tables are ubiquitous. There has been a variety of tasks that aim to automatically interpret, augment, and query tables. Current methods often require pretraining on tables or special model architecture design, are restricted to specific table types, or have simplifying assumptions about tables and tasks. This paper makes the first step towards developing open-source large language models (LLMs) as generalists for a diversity of table-based tasks. Towards that end, we construct \texttt{TableInstruct}, a new dataset with a variety of realistic tables and tasks, for instruction tuning and evaluating LLMs. We further develop the first open-source generalist model for tables, \texttt{TableLlama}, by fine-tuning Llama 2 (7B) with LongLoRA to address the long context challenge. We experiment under both in-domain setting and out-of-domain setting. 
On 7 out of 8 in-domain tasks, \texttt{TableLlama} achieves comparable or better performance than the SOTA for each task, despite the latter often has task-specific\nop{\hs{can we say `task-specific' as it'd be better}} design. On 6 out-of-domain datasets, it achieves 5-44 absolute point gains compared with the base model, showing that training on \texttt{TableInstruct} enhances the model's generalizability. We open source our dataset and trained model to boost future work on developing open generalist models for tables.\footnote{Code, model and data are available at: \url{https://osu-nlp-group.github.io/TableLlama/}.}
\end{abstract}
\section{Introduction}
\begin{figure*}[t]
\begin{center}
\includegraphics[width=0.99\linewidth]{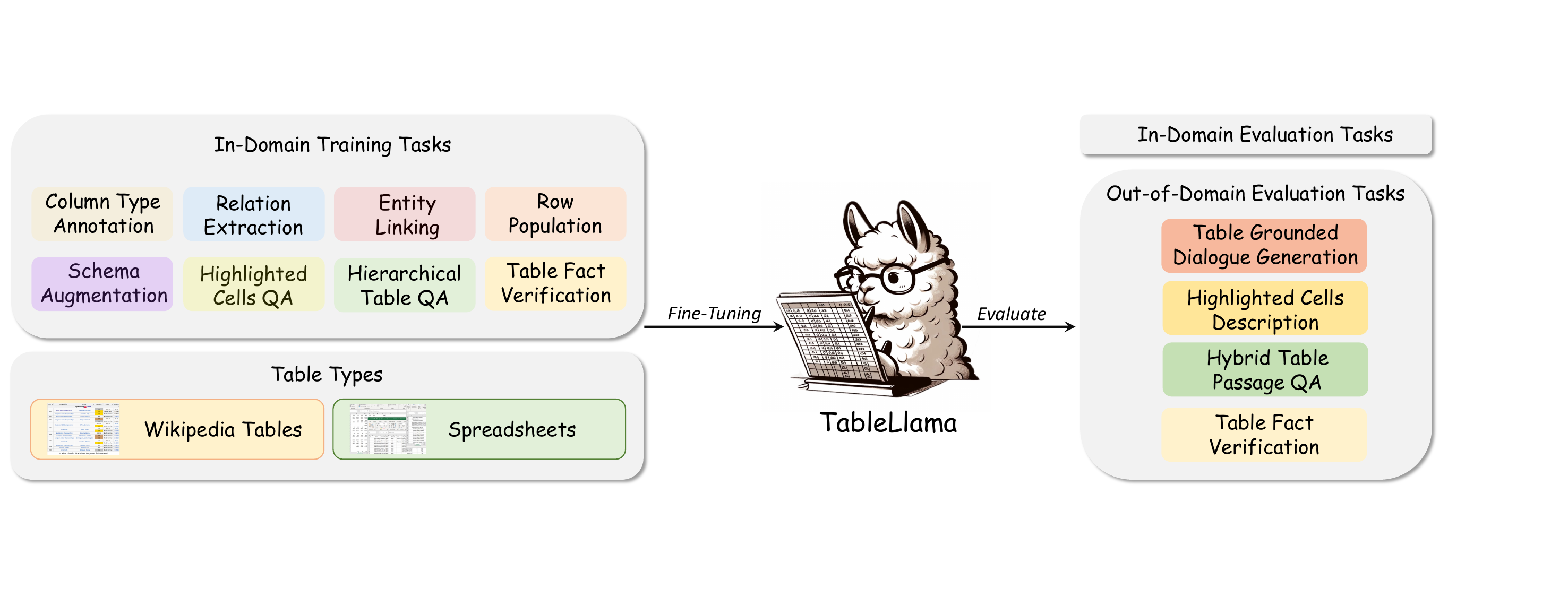}
\end{center}
\caption{An overview of \texttt{TableInstruct} and \texttt{TableLlama}. \texttt{TableInstruct} includes a wide variety of realistic tables and tasks with instructions. We make the first step towards developing open-source generalist models for tables with \texttt{TableInstruct} and \texttt{TableLlama}. \nop{The image of \texttt{TableLlama} is adapted based on DALL-E 3's generation.\hs{i think you could use the task name as the name for subfigure (b-d), and give a short description of what the task is under the table and then for each colored block, you change the name to sth like 'task instruction and expected output'? Also, font in the top subfigure should be changed; currently it's hard to read. do we need so many colors? I feel it's disturbing.}}}
\label{fig:overview}
\end{figure*}

Semi-structured tables are prevalent data structures to store and present information in almost every domain, ranging from scientific research, business reports, and healthcare records to financial statements.\nop{\hs{while it is fine to mention these domains, a question: tables in these domains usually have a lot of numbers, but in our datasets, we don't specifically deal with numbers, right? or some tasks actually have a lot of numbers?}}\nop{\zts{I feel it's ok. We have one hierarchical qa task that contains a lot of numbers}.} A variety of table-based tasks have been proposed, such as entity linking \citep{ritze2015matching},\nop{ column type annotation \citep{hulsebos2019sherlock},} schema augmentation \citep{zhang2017entitables}, and table-based question answering \citep{cheng-etal-2022-hitab, Nan2021FeTaQAFT,chen2020hybridqa}, which have spurred significant research interest \citep{deng2020turl, yin-etal-2020-tabert, wang2021tuta, iida-etal-2021-tabbie} in recent years. 

\nop{Building models for tables is not new.}
\nop{However, none of them meets all the requirements for a generalist model listed above.} 
Most existing methods for table-based tasks have at least one of the following limitations: (1) Require table pretraining \citep{liu2022tapex, yin-etal-2020-tabert, deng2020turl, iida-etal-2021-tabbie} and/or special model architecture design for tables \citep{deng2020turl, wang2021tuta, iida-etal-2021-tabbie}, (2) only support limited, specific types of tables and tasks \citep{Chen2020TabFact:, Nan2021FeTaQAFT}, (3) make strong simplifying assumptions (See the ``in-domain'' part of Section \ref{sec:simplifying-assum}) about tables and tasks \citep{li2023tablegpt}. 

On the other hand, language models like T5 \citep{raffel2020exploring} have been shown to excel in grounding language to structured knowledge \citep{xie2022unifiedskg}. In addition, instruction tuning \citep{chung2022scaling, wang-etal-2022-super, mishra-etal-2022-cross} appears as an important technique that can guide LLMs to follow instructions to complete a variety of tasks. 

\nop{\hs{it's too early to directly throw out this question.. in addtion to the above comment, you should also mention the recent trend showing the promise of using LLMs for multiple tasks via instruction tuning. therefore, you explore the following question. Currently there is too much a gap between your first sentence and this one.}} 
Under this background, we seek to answer the following question: \textit{Can we build a generalist model to handle a variety of table-based tasks using LLMs and instruction tuning?} Some exemplar tasks are shown in Figure \ref{fig:overview}.\nop{\zts{A generalist model for tables could ?
make the tables more easily to be manipulated and annotated, thus can serve as a user-friendly tool to largely reduce manual labor ().}}\nop{\hs{Are you trying to use the last sentence to motivate why building a generalist model vs fine-tuning a specific model for each task? if so, it's not convincing at all. why is it better than the latter in terms of "...more easily to be manipulated and annotated" and "as a user-friendly tool to largely reduce manual labor"?}} Such a generalist model shall meet the following desiderata: First, \textbf{it should not only work well on diverse table-based tasks, but also generalize to unseen tasks.} Since new table data and tasks can be constructed dynamically as new information arrives, it is hard to collect training data that covers all tasks and all tables, which requires a model to be inherently generalizable to tasks\nop{\hs{in our OOD setting, it has two sub-settings, right? one is to test on new tasks, the other to test on old tasks but new datasets}} and datasets it has never seen before. 
Second, \textbf{it should work on real-world tables and realistic tasks\nop{, which can be large, intricate, and incomplete}.} The model should not make strong assumptions to only handle simplified synthetic tables and tasks, but must embrace practical challenges such as handling complex numerical reasoning on large hierarchical spreadsheets as well as a large number of candidates for classification and ranking tasks. 

In pursuing this goal, we realize there lacks a comprehensive collection of realistic tables and tasks that can support the development and evaluation of generalist models. Therefore, we construct \texttt{TableInstruct}, by meticulously selecting representative table-based tasks from widely used datasets, unifying the format for all tasks and manually annotating instructions. \texttt{TableInstruct} shown in Table \ref{tab:benchmark} offers the following unique features: (1) \textbf{Diverse coverage of tables and tasks}. \texttt{TableInstruct} boasts a collection of 14 datasets of 11 tasks in total, with both in-domain and out-of-domain evaluation settings. Our training data includes 8 tasks, which are curated from 1.24M tables containing 2.6M instances spanning from table interpretation, table augmentation\nop{\hs{can you in table 1 mark what tasks are table interpretation and table augmentation? I don't think people are very familiar with these two.}}, table-based QA, and table-based fact verification. {We choose 8 datasets for these 8 tasks for in-domain evaluation and leave the other 6 datasets for 4 tasks for out-of-domain evaluation. The in-domain training tasks can enable the model to learn more fundamental table understanding abilities such as table interpretation and table augmentation, while we choose tasks that require more high-level reasoning abilities such as table QA and cell description to test the model's generalization ability.} This extensive range of tables and diverse tasks not only provide {valuable resources for table modeling}, but also foster a more comprehensive evaluation of generalist models. \nop{\hs{should we talk a bit more about the InD and OOD setting? how many/what are InD and OOD? Is there any rationale for choosing some as InD while others as OOD?}} (2) \textbf{The use of real-world tables and realistic tasks}. \texttt{TableInstruct} uses authentic real-world instead of overly simplified synthetic task data compared with existing work \citep{li2023tablegpt}. We incorporate a large number of Wikipedia tables and spreadsheets from statistical scientific reports\nop{, and collect\hs{I still have the earlier question: did we `collect'? if not, then reprase this as `as well as tables with varied length...'} tables} with varied length of contents, realistic and complex semantic types from Freebase \citep{freebase} for column type annotation and relation extraction, and a large referent entity corpus with rich metadata from Wikidata \citep{vrandevcic2014wikidata} for entity linking. In addition, we include complicated numerical reasoning tasks with hierarchical table structure and existing manually annotated \nop{\hs{do you mean they `manually annotated' or you did that?}}table QA and fact verification tasks. By doing so, we aim to equip models with the capability to cope with realistic and complex table-based tasks. 

\texttt{TableInstruct} requires models to accommodate long inputs\nop{\hs{can you in Table 1 show the min/max/median of the context length?}} (Table \ref{tab:benchmark}). We adopt LongLoRA \citep{longlora} based on Llama 2 (7B) \citep{touvron2023llama} as our backbone model, which has been shown efficient and effective to handle long contexts. We fine-tune it on \texttt{TableInstruct} and name our model \texttt{TableLlama}. We conducted extensive experiments\nop{ and analysis \hs{there is no analysis, right?}} under both in-domain and out-of-domain settings. Our experiments show \texttt{TableLlama} has strong capabilities for various in-domain table understanding and augmentation tasks, and also achieves promising performance in generalizing to unseen tasks and datasets. 

\begin{figure*}[t]
\begin{center}
\includegraphics[width=0.96\linewidth]{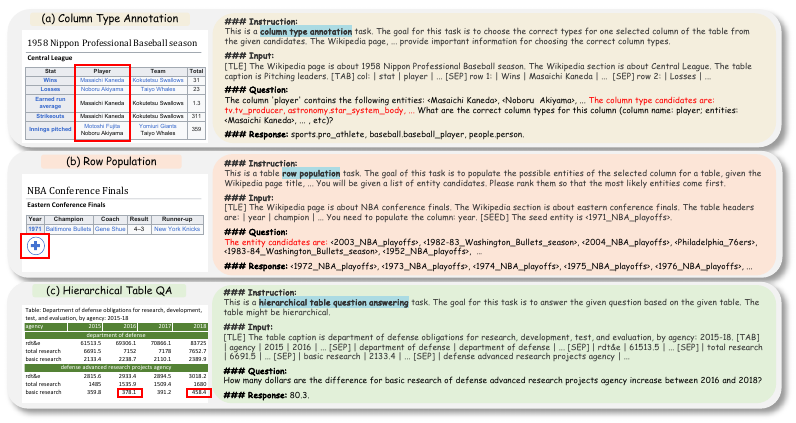}
\end{center}
\caption{Illustration of three exemplary tasks: (a) Column type annotation. This task is to annotate the selected column with the correct semantic types. (b) Row population. This task is to populate rows given table metadata and partial row entities. (c) Hierarchical table QA. For subfigures (a) and (b), we mark candidates with red color in the 
``task instruction'' part. The candidate set size can be hundreds to thousands in \texttt{TableInstruct}. \nop{\hs{1. can we remove the pink background of the TableLlama icon? 2. maybe separate subfigure (a) out as a standalone figure as Figure 1? those task formulations can be introduced later as Figure 2. BTW, I still do not like having so many colors in Figure (a).}}\nop{\hs{i think you could use the task name as the name for subfigure (b-d), and give a short description of what the task is under the table and then for each colored block, you change the name to sth like 'task instruction and expected output'? Also, font in the top subfigure should be changed; currently it's hard to read. do we need so many colors? I feel it's disturbing.}}}
\label{fig:three_examplars}
\end{figure*}

In summary, our main contributions are:
\begin{itemize}
    \item [$\bullet$] We construct \texttt{TableInstruct}, a large-scale instruction tuning dataset with diverse, realistic tasks based on real-world tables. We unify their format and manually annotate instructions to guarantee quality. 
    \item [$\bullet$] We develop \texttt{TableLlama}, an open-source LLM-based generalist model fine-tuned on \texttt{TableInstruct}. Experiments show that compared with the SOTA on each task that often has special pre-training or model architecture design for tables, \texttt{TableLlama} can achieve similar or even better performance on almost all of the in-domain tasks\nop{\hs{it is confusing here. what do you mean by this: when using the same training data for tables}}. For out-of-domain tasks, compared with the base model, \texttt{TableLlama} can achieve 5-44 absolute point gains on 6 datasets, and compared with GPT-4, \texttt{TableLlama} has less gap or even better zero-shot performance on 4 out of 6 datasets, which demonstrate that \texttt{TableInstruct} can substantially enhance model generalizability. 
    \nop{\hs{I think the last part isn't very convincing; why comparing with pre-trained model? I think you can admit for OOD, there is still some gap with SoTAs, which is understandable, and then stress on the improvement over the pre-trained/un-fine-tuned model}}
\end{itemize}

\nop{\hs{In this paragraph, you should talk about your effort to unify the task format and prepare instructions (briefly), followed by what models you trained to do instruction tuning, as well as how you solved the long context issue.}}
\nop{\hs{Give a summary of the results in this paragraph and maybe also talk about insights for future work.}}


\section{\texttt{TableInstruct} Benchmark}

\begin{table*}[]
\small
\centering
\resizebox{\linewidth}{!}{
\begin{tabular}{@{}lllcccccc@{}}
\toprule
\multirow{2}{*}{Task Category} &\multirow{2}{*}{Task Name} & \multirow{2}{*}{\quad\quad\quad Dataset} & \multirow{2}{*}{
\begin{tabular}[c]{@{}c@{}}In-\\ domain\end{tabular}
} & \#Train & \#Test & \multicolumn{3}{c}{Input Token Length}\\ 
& & & & (Table/Sample) & (Table/Sample) & min & max & median \\
\midrule
\multirow{3}{*}{  
\begin{tabular}[c]{@{}l@{}}Table\\ Interpretation\end{tabular}
} &Col Type Annot. & 
\multirow{3}{*}{TURL~\citep{deng2020turl}} & Yes & 397K/628K & 1K/2K & 106 & 8192 & 2613\\ 
&Relation Extract. & & Yes & 53K/63K & 1K/2K & 2602 & 8192 & 3219\\ 
&Entity Linking & & Yes & 193K/1264K & 1K/2K & 299 & 8192 & 4667\\ \midrule
\multirow{2}{*}{ 
\begin{tabular}[c]{@{}l@{}}Table\\ Augmentation\end{tabular}
} & Schema Aug. & \multirow{2}{*}{TURL~\citep{deng2020turl}} & Yes & 288K/288K & 4K/4K & 160 & 1188 &215\\ 
& Row Pop. &  & Yes & 286K/286K & 0.3K/0.3K & 264 & 8192 & 1508\\ \midrule
\multirow{5}{*}{ 
\begin{tabular}[c]{@{}l@{}}Question\\ Answering\end{tabular}
} & Hierarchical Table QA &HiTab~\citep{cheng-etal-2022-hitab} & Yes & 3K/7K & 1K/1K & 206 &5616 & 978\\
& Highlighted Cells QA &FeTaQA~\citep{Nan2021FeTaQAFT} & Yes & 7K/7K & 2K/2K & 261 & 5923 & 740\\ 
&Hybrid Table QA & HybridQA~\citep{chen2020hybridqa} & No & \text{--} & 3K/3K & 248 & 2497 &675\\
&Table QA&WikiSQL~\citep{wikisql} & No & \text{--} & 5K/16K & 198 & 2091 & 575\\
&Table QA&WikiTQ~\citep{wikitq} & No & \text{--} & 0.4K/4K & 263 & 2688 & 709\\
\midrule
\multirow{2}{*}{ 
\begin{tabular}[c]{@{}l@{}}Fact\\ Verification\end{tabular}
} & \multirow{2}{*}{Fact Verification} & TabFact~\citep{Chen2020TabFact:} & Yes & 16K/92K & 2K/12K & 253 & 4975 & 630\\ 
&& FEVEROUS~\citep{feverous} & No & \text{--} & 4K/7K & 247 & 8192 & 648\\
\midrule
\begin{tabular}[c]{@{}l@{}}Dialogue\\ Generation\end{tabular}
  & \begin{tabular}[c]{@{}l@{}}Table Grounded \\ Dialogue Generation\end{tabular} & KVRET~\citep{kvret} & No & \text{--} & 0.3K/0.8K & 187 & 1103 & 527\\
\midrule
Data-to-Text &  
\begin{tabular}[c]{@{}l@{}}Highlighted\\ Cells Description \end{tabular}
& ToTTo~\citep{parikh-etal-2020-totto} & No & \text{--} & 7K/8K & 152 & 8192 & 246\\
\bottomrule
\end{tabular}
}
\caption{Statistics of train/test tasks and datasets in our \texttt{TableInstruct}. For each task, we explain its definition and show an example in Appendix \ref{sec:prompt_format}. \nop{\zts{TODO: add context length} \hs{why is there a `-' for `Table'? does it mean only 1 table? if so, just put 1? is `conversational' the full name?} \hs{For each task, we explain its definition and show an example in Appendix XX.}}}
\label{tab:benchmark}
\end{table*}

Unlike existing datasets predominantly designed for training task-specific table models, our objective is to bridge the gap between multiple complex task-specific models and one simple generalist model that can deal with all the table-based tasks without extra model-design efforts. To achieve this, our approach for constructing \texttt{TableInstruct} adheres to the following principles. First, instead of collecting multiple datasets from highly homogeneous tasks, we try to diversify the tasks and table types. We pick representative table-based tasks that necessitate different abilities of models, such as table interpretation, table augmentation, table QA and table fact verification from Wikipedia tables and spreadsheets in statistical scientific reports. Second, we select realistic tasks and construct high-quality instruction data in a unified fashion without simplifying assumptions (see ``in-domain'' part of \ref{sec:simplifying-assum})\nop{\lyf{I think "simplifying assumptions" this term occurs many times so far. We should clarify this concept at the first time}}. \texttt{TableInstruct} will support {powerful modeling and realistic evaluation approaches, ensuring a valuable and practical dataset for research.

\subsection{Data Collection}

\texttt{TableInstruct} incorporates samples from 14 table-based datasets of 11 distinctive tasks (Table \ref{tab:benchmark}). We separate them and select 8 datasets of 8 tasks for training and in-domain evaluation. We leave the other 6 datasets of 4 tasks as held-out unseen datasets for out-of-domain evaluation. 

\noindent \textbf{Task category:} Tasks in \texttt{TableInstruct} can be categorized into several groups: table interpretation, table augmentation, question answering, fact verification, dialogue generation, and data-to-text. Table interpretation aims to uncover the
semantic attributes of the data contained in relational tables, and
transform this information into machine understandable knowledge. Table augmentation is to expand the partial tables with additional data. Question answering aims to obtain the answer with tables and optional highlighted cells or passages as evidence. Fact verification is to discriminate whether the tables can support or refute the claims. Dialogue generation is to generate a response grounded on the table and dialogue history. Data-to-text is to generate a description based on the highlighted cells.\nop{We choose table interpretation, table augmentation, some representative question answering tasks and one dataset of fact verification as in-domain datasets for training the model. We hold out the rest of the datasets spanning from question answering, fact verification, dialogue generation to data-to-text as out-of-domain datasets.} By choosing the tasks that require models to learn more fundamental table understanding abilities such as table interpretation and table augmentation for training, we hope the model can demonstrate generalization ability on out-of-domain datasets such as high-level table QA and table cell description tasks.
\nop{\hs{maybe here you briefly introduce the task category like what table interpretation in general does, and then introduce your rationale why some are used for in-domain evaluation and some are for ood.}}

\noindent\textbf{In-domain:} \label{sec:simplifying-assum} 
The tasks for training the generalist table model include column type annotation, relation extraction, entity linking, row population, schema augmentation, hierarchical table QA, highlighted cells QA, and table fact verification.\nop{The tasks for training the generalist table model include column type annotation \citep{deng2020turl}, relation extraction \citep{deng2020turl}, entity linking \citep{deng2020turl}, row population \citep{deng2020turl}, schema augmentation \citep{deng2020turl}, hierarchical table QA \citep{cheng-etal-2022-hitab}, highlighted cells QA \citep{Nan2021FeTaQAFT}, and table fact verification \citep{Chen2020TabFact:}.} These tasks require the model to understand the semantics of table columns, the relation between table column pairs, the semantics of table cells and require the model to gain reasoning ability to answer table-related questions and verify the facts. For the dataset of each task, we intentionally pick up those that enjoy realistic task complexity without simplifying assumptions. For example, for column type annotation and relation extraction, these two tasks are multi-choice classification tasks in essence. We use real-world column semantic types and relation types from Freebase \citep{freebase}, which contains hundreds of complex choices such as ``government.politician.party\-government.political\_party\_tenure.party'' shown in Figure \ref{fig:rel_extraction} in Appendix \ref{sec:prompt_format}. For entity linking, the referent entities are from real-world Wikidata \citep{vrandevcic2014wikidata}, which contains hundreds of complex metadata, such as ``<2011-12 Melbourne Victory season
[DESCRIPTION] Association football club 2011/12 season for Melbourne Victory [TYPE]
SoccerClubSeason>'' as shown in Figure \ref{fig:ent_link} in Appendix \ref{sec:prompt_format}. For schema augmentation and row population, there are a huge number of candidates that LLMs need to rank. For hierarchical table QA, all the tables are engaged with intricate structures with multi-level column names and row names. In addition, it is intensive in numerical reasoning which requires LLMs to understand table structure, identify related cells and do calculations. By doing so, we hope to enable LLMs to become truly powerful generalist models that can handle sophisticated table tasks and \texttt{TableInstruct} can be a realistic benchmark to evaluate LLMs’ abilities compared with specially designed table models.

\noindent\textbf{Out-of-domain:} A powerful generalist table model is expected to not only demonstrate strong performance on in-domain tasks, but also generalize well to unseen tasks or unseen datasets of the same tasks. \nop{The underlying table understanding ability learned by the model should be able to transfer to unseen tasks or datasets.} We choose tasks such as table QA and cell description that require the model's high-level table understanding and reasoning ability as out-of-domain datasets. We involve HybridQA \citep{chen2020hybridqa}, KVRET \citep{kvret}, FEVEROUS \citep{feverous}, ToTTo \citep{parikh-etal-2020-totto}, WikiSQL \citep{wikisql} and WikiTQ \citep{wikitq} as 6 out-of-domain datasets to test our model's generalization ability.
\nop{HybridQA is a table and passages grounded question answering task. KVRET is a response generation task grounded on table and dialogue history. ToTTo is to generate text descriptions based on highlighted table cells. FEVEROUS is a table fact verification task. WikiSQL and WikiTQ are two table QA tasks\hs{you don't have such a description for each task in the previous paragraph. Maybe as said in the caption of Table 1" leave such a brief description in appendix.}. By evaluating our model on these datasets, we hope to demonstrate our model's generalization ability. }
\nop{\hs{provide the rationale why some are in-domain while others are OOD. polish the language you said on Teams}}

\subsection{Task Formulation and Challenges}
\nop{\hs{I think you should first talk about Section 3.2 (what these tasks are) and then talk about Section 3.1 (how to formulate them into a unified format). Do point readers to some examples, whether in the main content or in the appendix. Otherwise, it's very boring to read.}}

\nop{\hs{this paragraph could be significantly polished. you could refer to Figure \ref{fig:examplars} b-d in the beginning of this paragraph when talking about the prompt format. and then stress on the challenges in our task formulation, e.g., number of candidates and context length.}} The primary objective of \texttt{TableInstruct} is to design one generalist model for all table-based tasks. As Figure \ref{fig:three_examplars} (a)-(c) shows, each instance in our dataset maps three components: <instruction, table input, question> to an output. The instruction is manually designed to point out the task and give a detailed task description. We concatenate table metadata such as the Wikipedia page title, section title and table caption with the serialized table as table input. In the question, we put all the information the model needed to complete the task and prompt the model to generate an answer. For example, for the column type annotation task, as Figure \ref{fig:three_examplars} (a) shows, the column named ``Player'' needs to be annotated with its semantic types. In the format, the ``instruction'' gives the description of the task. The ``input'' contains the table-related information. Then we provide the entire candidate pool in the ``question'' and ask the model to choose one or multiple correct semantic types for this column. 

\noindent \textbf{Challenges.} Since we select realistic tasks and tables, the table length can vary from several to thousands of rows. Besides, for some tasks that are essentially multi-choice classification or ranking, the entire candidate pool can be very large up to thousands. Furthermore, as the candidates are from real-world Freebase \nop{\citep{freebase}}and Wikidata\nop{\citep{vrandevcic2014wikidata}}, each candidate is long, such as ``<2011-12 Melbourne Victory season [DESCRIPTION] Association football club 2011/12 season for Melbourne
Victory [TYPE] SoccerClubSeason>'' is one candidate for entity linking. These characteristics can not only make it difficult for the model to learn, but also introduce the challenge of handling long contexts.

\nop{\hs{maybe allocate one specific paragraph to talk about challenges of our task formulation. for example, move some stuff in the above paragraph and the first few sentences in the `model selection' paraphraph in Section 3 to here.}}



\section{Experimental Setup}

\begin{table*}[]
\small
\centering

\begin{tabular}{@{}llcclcc@{}}
\toprule
\multicolumn{7}{c}{In-domain Evaluation} \\
\midrule
 Datasets & Metric & Base & TableLlama & \qquad \qquad SOTA & GPT-3.5 & GPT-4\S \\ 
 \midrule
Column Type Annotation &  F1 & 3.01 & 94.39 & \textbf{94.54}*\dag~\citep{deng2020turl} & 30.88 & 31.75\\
Relation Extraction & F1 & 0.96 & 91.95 & \textbf{94.91}*\dag~\citep{deng2020turl} & 27.42 & 52.95 \\
Entity Linking & Accuracy & 31.80 & \textbf{93.65} & 84.90*\dag~\citep{deng2020turl} & 72.15 & 90.80\\ 
Schema Augmentation & MAP & 36.75 & \textbf{80.50} & 77.55*\dag~\citep{deng2020turl} & 49.11 & 58.19\\ 
Row Population & MAP & 4.53 & 58.44 & \textbf{73.31}*\dag~\citep{deng2020turl} & 22.36 & 53.40 \\ 
HiTab & Exec Acc & 14.96 & \textbf{64.71} & 47.00*\dag~\citep{cheng2021fortap} & 43.62 & 48.40\\
FeTaQA & BLEU & 8.54 & \textbf{39.05} & 33.44~~~~~\citep{xie2022unifiedskg} & 26.49 & 21.70\\ 

TabFact & Accuracy & 41.65 & 82.55 & \textbf{84.87}*~~~\citep{zhao-yang-2022-table} & 67.41 & 74.40\\ 

\bottomrule
\end{tabular}
\caption{In-domain evaluation results. ``Base'': LongLoRA model w/o fine-tuning on \texttt{TableInstruct}; ``*'': w/ special model architecture design for tables/tasks; ``\dag'': w/ table pretraining; ``\S": for GPT-4, we uniformly sample 500 examples from test set for each task due to limited budget.\nop{For column type annotation and entity linking, we uniformly sample a subset from the original test data as our test set due to the large test size. For row population, we filter out the examples with more than 500 candidate entities from the original test set and randomly sample a subset as our test set. For all the downsampled test set, we reproduce the SOTA results using the SOTA model.} \nop{\zts{For column type annotation and relation extraction, as the candidate size is large, the base model tends to continue generating different candidates without stopping, which shows the model doesn't have the ability to choose the correct answer from given candidates for these two tasks.} \nop{\hs{you should comment a bit on those extremely low numbers like why it achieves zeros.}}}}
\label{tab:main_results_in_domain}
\end{table*}

\noindent\textbf{Model Construction.} \nop{As we intentionally select realistic tasks and real-world tables, the table length can vary from several to thousands of rows. Besides, for some tasks that are basically multi-choice classification or ranking, the entire candidate pool can be very large up to hundreds or thousands. \hs{move these sentences to Section 2.2: challenges?} Furthermore, as the candidates are from real-world Freebase \citep{freebase} and Wikidata \citep{vrandevcic2014wikidata}, each candidate is long. These characteristics can not only increase the difficulty for the model to learn, but also introduce the challenge for the model to handle the long input context. Although a few existing LLMs \citep{chen2023extendingcontextllm32k, tworkowski2023focusedtransformer} can handle more than 4K context, their training time is quadratically increasing with context length, which becomes very costly for us to further fine-tune them on \texttt{TableInstruct} due to our large data scale.} Although a few existing LLMs \citep{chen2023extendingcontextllm32k, tworkowski2023focusedtransformer} can handle longer than 4K contexts, their training time is quadratically increasing with context length, which becomes very costly for us to further fine-tune them on \texttt{TableInstruct} due to our large data scale. As LongLoRA \citep{longlora} has been shown as an effective and efficient technique to train long-context LLMs with shift short attention, we adopt it as our backbone model. Shift short attention splits context length into several groups and conducts attention in each group individually. The
tokens are shifted by half group size in half attention heads to ensure the information flow between neighboring groups. For example, LongLoRA can use shift short attention with group size 2048 to approximate total 8196 context length training, which leads to less computation cost with similar performance compared to fine-tuning with vanilla attention.
\nop{\hs{it's better to mention the base LLM LongLoRA is based on and a bit more details on how it enables long context training, as it is a pretty new tech.}}We fine-tune LongLoRA on \texttt{TableInstruct} to get our generalist model \texttt{TableLlama}.

\nop{
\noindent\textbf{Datasets.}
\hs{this part seems to be highly duplicated with Section 2.1. Do we still need it here?} We have selected 14 in-domain and out-of-domain datasets (Table \ref{tab:benchmakr}) across different tasks for tables. For in-domain datasets, we consider datasets curated by TURL \citep{deng2020turl} for five tasks: column type annotation, relation extraction, entity linking, schema augmentation and row population. We utilize FeTaQA \citep{Nan2021FeTaQAFT} for free-form table QA task, TabFact \citep{Chen2020TabFact:} for table fact verification task, HiTab \citep{cheng-etal-2022-hitab} for hierarchical table QA task. For out-of-domain datasets, we use FEVEROUS \citep{feverous} for table fact verification task, HyrbidQA \citep{chen2020hybridqa} for table and passages grounded QA task, KVRET \citep{kvret} for table and dialogue history grounded response generation task, ToTTo \citep{parikh-etal-2020-totto} for highlighted cells description task, WikiSQL \citep{wikisql} and WikiTQ \citep{wikitq} for table QA tasks.}

\noindent\textbf{Existing SOTA Models}. In our evaluation settings, we have 9 out of 14 SOTA models utilize table pretraining and/or have special model architecture design for tables. The detailed description for each SOTA model is in Appendix \ref{sota_model}.

\nop{TURL \citep{deng2020turl} is an encoder-based BERT-like model pre-trained on 570K tables. Though TURL has shown SOTA performance on various table tasks such as column type annotation, relation extraction, entity linking, row population and schema augmentation, it requires fine-tuning task-specific modules on labeled data.\nop{when generalizing to new tasks or new datasets} The SOTA method for HiTab builds on 1) TUTA \citep{wang2021tuta}, which uses tree attention as the encoder to capture table structures and 2) FORTAP \citep{cheng2021fortap}, which leverages
spreadsheet formulas for table pre-training to better handle numerical reasoning. The SOTA method for TabFact designs a self-labeled keypoint alignment \citep{zhao-yang-2022-table} to align salient evidence and aggregate essential information between the statement and table. For HybridQA, the SOTA method MATE \citep{eisenschlos2021mate} uses sparse attention for Transformer architecture which allows heads to efficiently attend to either rows
or columns in a table. The SOTA method for WikiSQL and WikiTQ is TAPEX \citep{liu2022tapex}, which fuses table pre-training by learning a neural SQL executor over a synthetic corpus. For FeTaQA, FEVEROUs, KVRET and ToTTo, the SOTA results come from T5-3B fine-tuned on their own individual training data \citep{xie2022unifiedskg}. 
\nop{\hs{is this from unifiedSKG? why there is no citation?}. }}

\nop{\hs{[I think it is worth introducing the baselines, which achieve the SoTAs. otherwise it is not clear the advantage of our approach to them]}}

\noindent\textbf{Evaluation Metrics.}
We follow the above baselines to use their evaluation metrics.\nop{\lyf{use the same evaluation metrics as them / keep evaluation metrics the same?}} For column type annotation, relation extraction and KVRET, we use Micro F1. For entity linking, TabFact, FEVEROUS, HybridQA, WikiSQL and WikiTQ, we use accuracy. For row population and schema augmentation, we use MAP. For Hitab, we use execution accuracy \citep{wikisql}. For FeTaQA and ToTTo, we use BLEU \citep{papineni-etal-2002-bleu}.

\noindent\textbf{Training and Inference Details.}
We choose LongLoRA 7B \citep{longlora}, fully fine-tuning version with 8K context length limit as our base model. The fully fine-tuning version replaces the vanilla attention in Llama 2 with shift short attention. We fine-tune the model with Huggingface transformers library \citep{wolf-etal-2020-transformers}. We merge all eight datasets and repeat three smaller datasets (i.e., FeTaQA, HiTab and TabFact) for six times and randomly shuffle them as our final training data. We use a learning rate of 2e-5 and set the batch size at 3. We streamingly train the model on 48 A100 80GB GPUs and use a cosine scheduler with a 3\% warm-up period for 2 epochs\nop{, which lasts for 9 days}. To efficiently train the model, we employ DeepSpeed training with ZeRO-2 stage \citep{rajbhandari2020zero_deepspeed}. For both training and inference, we set the input length as 8192. For inference on \texttt{TableLlama}, as different tasks have different lengths of the ground truth, we use 64 as the output length for column type annotation, relation extraction, entity linking, HiTab, TabFact, FEVEROUS, HybridQA, WikiSQL and WikiTQ, 128 for schema augmentation, FeTaQA, KVRET and ToTTo, and 512 for row population. For column type annotation and entity linking, we uniformly sample a subset from the original test data as our test set due to the large test size. For row population, we filter out the examples with more than 500 candidate entities from the original test set and randomly sample a subset as our test set. For all the downsampled test set, we reproduce the SOTA results using the SOTA model.

For closed-source LLMs, we use the gpt-4-1106-preview version for GPT-4, which is the latest version that supports 128K context and reports the best performance. For GPT-3.5, we use the gpt-3.5-turbo-1106 version, which supports 16K context.

\section{Result Analysis}

\begin{table*}[t]
\small
\centering
\begin{tabular}{@{}llcclccc@{}}
\toprule
\multicolumn{8}{c}{Out-of-domain Evaluation} \\
\midrule
 Datasets & Metric & Base & TableLlama & \qquad \qquad SOTA & $\Delta_{Base}$  & GPT-3.5 & GPT-4\S\\
 \midrule
FEVEROUS & Accuracy & 29.68 & 73.77 & \color{gray}85.60~~~\citep{tay2022ul2} & +44.09 & 60.79 & 71.60 \\
HybridQA & Accuracy & 23.46 & 39.38 & \color{gray}65.40*~\citep{lee-etal-2023-mafid} & +15.92 & 40.22 & 58.60\\
KVRET & Micro F1 & 38.90 & 48.73 & \color{gray}67.80~~~\citep{xie2022unifiedskg} & +9.83 & 54.56 & 56.46\\
ToTTo & BLEU & 10.39 &20.77 & \color{gray}48.95~~~\citep{xie2022unifiedskg} & +10.38  & 16.81 & 12.21 \\
WikiSQL & Accuracy & 15.56 & 50.48 & \color{gray}92.70~~~\citep{xu2023sead} &+34.92 & 41.91 & 47.60\\
WikiTQ & Accuracy & 29.26 & 35.01 & \color{gray}57.50\dag~\citep{liu2022tapex} & +5.75 & 53.13 & 68.40\\
\bottomrule
\end{tabular}
\caption{Out-of-domain evaluation results. ``Base'': LongLoRA model w/o fine-tuning on \texttt{TableInstruct}; ``*'': w/ special model architecture design for tables/tasks; ``\dag'': w/ table pretraining; ``\S": for GPT-4, we uniformly sample 500 examples from test set for each task due to limited budget. We put the SOTA performances here in grey for reference and note that they were achieved under full-dataset training for each task while \texttt{TableLlama} is zero-shot. \nop{\zts{The SOTA results come from full-dataset training while the results for \texttt{TableLlama} are zero-shot, so it's unfair to make a direct comparison between them. We put the SOTA results here for reference.}} \nop{\hs{can we make the SOTA's number in grey, basically put them here and try not to let people focus on them. you should also point out the SOTA are simply for referrence and it is unfair to make a direct comparison.}}}
\label{tab:main_results_out_domain}
\end{table*}

\subsection{Main Results}
\noindent\textbf{In-domain Results.} As Table \ref{tab:main_results_in_domain} shows, we train \texttt{TableLlama} on eight table-based tasks and evaluate it on their test sets as the in-domain results. Due to the special semi-structured nature of tables, for most table-based tasks, existing work achieves SOTA results by using pretraining on large-scale tables and/or special model architecture design tailored for tables. Nonetheless, we observe that: 

\textit{By simply fine-tuning a large language model on \texttt{TableInstruct}, \texttt{TableLlama} can achieve comparable or even better performance on almost all the tasks without any table pretraining or special table model architecture design.} \nop{\hs{people likely are not familiar with how the SoTA models works. you need to explain that and then say the previous sentence. otherwise, they cannot get what you're saying.}} For most of the tasks, the performance gap is within 3 absolute points, except for row population. For entity linking, schema augmentation, HiTab and FeTaQA, \texttt{TableLlama} can exceed the SOTA performance by up to 17.71 absolute points. This demonstrates that empowering open-source LLMs with more powerful table understanding abilities via instruction tuning can be a promising research direction to further explore.

\textit{\texttt{TableLlama} displays advantanges\nop{\hs{I wouldn't call it `proficiency' due to the still low numbers. maybe `advantage'?}} in table QA tasks.} HiTab and FeTaQA are two table question answering tasks we include for training. By comparing the results, we found that \texttt{TableLlama} can surpass the SOTA by 5.61 points for FeTaQA and 17.71 points for HiTab, which is full of numerical reasoning on tables. As LLMs have been shown superior in interacting with humans and answering questions, this indicates that\nop{\hs{does this make sense? do you mean those SOTA do not have language understanding ability?}} the existing underlying strong language understanding ability of LLMs may be beneficial for such table QA tasks despite with semi-structured tables.

For entity linking which requires the model to link the mention in a table cell to the correct referent entity in Wikidata, \texttt{TableLlama} also presents superior performance with 8 points gain over SOTA. Since the candidates are composed of referent entity name and description, we hypothesize LLMs have certain abilities to understand the description which help identify the correct entities.

Row population is the only task that \texttt{TableLlama} has a large performance gap compared to the SOTA. Here we provide a large number of candidates for the model to rank given table metadata and the seed row entity. By analyzing the errors, we found that the model can easily identify the entities containing similar numbers in sequence, such as the first example shown in Table \ref{tab:row_pop} in Appendix \ref{sec:case_study}. However, for entities that share high similarities\nop{\hs{what does this `high similarities for some attributes' mean?}}, such as the second example in Table \ref{tab:row_pop} shows, the target row entities are the competitions which ``Oleg Veretelnikov'' got achievements in. To correctly populate the entities from the given plenty of candidates highly related to ``competitions'', it requires the model to understand the inherent relation between the athlete and each given candidate\nop{their previously attended competitions}, which is still challenging for the current model.

\nop{belonging to the same attributes but different names, for example, if the target entities are the members of a team, by giving different names which come from irrelevant teams, it's hard for the model to distinguish which one belongs to the required team.\hs{hard to understanding...can you provide some examples in appendix or show an example or two in main content? This part feels more rushed.}}

\noindent\textbf{Out-of-domain results.} We evaluate \texttt{TableLlama} on six out-of-domain datasets. We observe that:

\textit{By comparing with the base model, \texttt{TableLlama} can achieve 5-44 points gain on 6 out-of-domain datasets, which demonstrates \texttt{TableInstruct} can enhance the model's generalization ability.} By learning from the table-based training tasks, the model has acquired essential underlying table understanding ability, which can be transferred to other table-based tasks/datasets and facilitate their performance. Among these 6 datasets, we found that FEVEROUS, a table fact verification dataset exhibits the largest gain over the other 5 datasets. This is likely because the fact verification task is an in-domain training task, despite the dataset unseen during training. Compared with cross-task generalization, it may be easier to generalize to different datasets belonging to the same tasks. \nop{\zts{In addition, comparing the model trained on TabFact and \texttt{TableInstruct}, when evaluating on FEVEROUS, which is the same task transfer for TabFact, we found \texttt{TableLlama} can achieve 72.30 accuracy while the model trained on TabFact can only achieve 56.15 accuracy. This indicates that other tasks in the training set that don't share the same task with FEVEROUS also play an important role in engaging the model to obtain stronger table fact verification ability.}

\zts{\textit{The model trained on table QA tasks has better generalization ability than the model trained on other tasks.} As Table \ref{tab:out_domain_ablation} shows, the model trained on HiTab and FeTaQA can achieve 28.92 and 25.03 overall performance respectively, which surpasses the performance trained on the other 6 tasks in a large gain. However, including those tasks (i.e., \texttt{TableInstruct}) to train the model can further enhance the model's generalization ability, which achieves 40.45 overall performance on 6 out-of-domain datasets.} \hs{this could be moved to the new Section 4.2?}}

Although there is still some gap between our performance and the previously reported SOTA for each dataset, we note those SOTAs were achieved under full-dataset training while \texttt{TableLlama} is zero-shot, hence it is reasonable to see such a gap. Nevertheless, we hope our work can inspire future work to further improve the zero-shot performance. 

\noindent\textbf{Open-source vs. closed-source.} We compare \texttt{TableLlama} and closed-source LLMs (i.e., GPT-3.5 and GPT-4) and observe that:

\textit{\texttt{TableLlama} achieves better performance on in-domain tasks compared with closed-source LLMs.} It shows that even if closed-source LLMs have demonstrated strong performance in general, fine-tuning open-source LLMs on task-specific table-based data still has better performance. 

\textit{\texttt{TableLlama} shows less gap or even better zero-shot performance than closed-source LLMs on 4 out of 6 out-of-domain datasets (i.e., FEVEROUS, KVRET, ToTTo and WikiSQL), which shows TableLlama has gained generalization ability.} But closed-source LLMs are still stronger at table-based QA tasks that require more complex reasoning.

GPT-4 has better results than GPT-3.5 on all the in-domain and out-of-domain datasets except for FeTaQA and ToTTo. This is because GPT-4 generates longer output than GPT-3.5, so for FeTaQA and ToTTo which are evaluated using BLEU to compare the generated sentence the ground truth sentence, GPT-3.5 performs better.

\subsection{Ablation Study}
\begin{table*}[t]
\centering
\setlength{\tabcolsep}{1mm}
\resizebox{\linewidth}{!}{
\begin{tabular}{@{}lccccccccccccccc@{}}
\toprule
 \multirow{3}{*}{\begin{tabular}[c]{@{}l@{}}Training \\ Data\end{tabular}} & \multicolumn{8}{c}{In-domain}& \multicolumn{6}{c}{Out-of-domain} \\
 \cmidrule(l){2-9}
 \cmidrule(l){10-15}& ColType & RelExtra & EntLink & ScheAug & RowPop & HiTab & FeTaQA & TabFact & FEVER. & HybridQA & KVRET & ToTTo & WikiSQL & WikiTQ\\ \cmidrule(l){2-15} 
 & F1 & F1 & Acc & MAP & MAP & Acc & BLEU & Acc & Acc & Acc & Micro F1 & BLEU & Acc & Acc\\ \midrule
Base & 3.01 & 0.96 & 31.80 & 36.75 & 4.53 & 14.96 & 8.54 &41.65 &  29.68& 23.46 & 38.90 & 10.39 & 15.56 & 29.26\\ \midrule
ColType & 94.32 & 0 & 0 & 0 & 0 & 0.13 & 0.52 & 0  & 0 & 0 & 0 & 1.11 &0.35 & 0.21\\
 RelExtra & 45.69& \textbf{93.96} & 0.45 & 8.72 & 0.99& 7.26 & 1.44 & 0 & 2.38 & 8.17 & 5.90 & 5.60 & 7.02 & 9.58\\
 EntLink &0.86 & 0.03 & 88.45 & 2.31 & 0.94 &5.37 & 4.79 & 0  & 39.04 & 3.06 & 0 & 1.76 & 3.42 & 7.07 \\
 ScheAug  & - & - & - & 80.00 & - & - & - & - & - & - & - & - & - & - \\
RowPop  & - & - & - & - & 53.86 & - & - & -& - & - & - & - & - & -  \\ HiTab & 0.20 & 0.14 & 7.15 & 40.81 & 5.45 & 63.19 & 2.07 & 49.46  & 46.81 & 24.70 & 38.70 & 2.45 & 32.86 & 27.97  \\
FeTaQA & 0 & 0.40 & 0 & 30.23 & 0.15 & 19.57 & 38.69 & 1.20  & 1.21 & 33.79 & \textbf{50.69} & \textbf{23.57} & 13.79 & 27.12\\
TabFact & 0 & 0 & 0 & 0 & 0 & 0 & 0 & 74.87 & 56.15 & 0 & 0 & 0 & 0 & 0\\ \midrule
 \textbf{TableInstruct}   & \textbf{94.39} & 91.95 & \textbf{93.65} & \textbf{80.50} & \textbf{58.44} & \textbf{64.71} &\textbf{39.05} & \textbf{82.55}  & \textbf{73.77} & \textbf{39.38} & 48.73 & 20.77 & \textbf{50.48} & \textbf{35.01}\\  \bottomrule
\end{tabular}%
}
\caption{Transfer between different datasets. Bold numbers are the best results for each evaluation dataset. For models trained on schema augmentation (ScheAug) and row population (RowPop), their predictions on other datasets tend to repeat the candidates in the training data, which means they cannot generalize to other datasets, and hence we use ``-'' to represent their performances.}
\label{tab:entire_ablation}
\end{table*}

To better understand how \texttt{TableInstruct} helps enhance the model's generalizability, we conduct an ablation study to show the transfer between individual datasets.


\textit{The model trained on table-based QA tasks\nop{\hs{you have to be precise.. in Table 1, previously you use table QA to refer to WikiSQL and WikiTQ..}} generalizes better than that trained on other tasks.} \nop{\hs{I think in this section, we should talk about the transfer performance between different task families (especially since earlier we mentioned `task families'). It is not surprising that a model transfer better to different datasets under the same task (although it is still good to discuss this.) }}As Table \ref{tab:entire_ablation} shows, the model trained on HiTab scores more than 20 points on 7 out of 13 unseen datasets\nop{\hs{why is this? this is because those 7 datasets are similar to the training set, right, although they might not be the same task.}}, and that trained on FeTaQA scores more than 10 points on 7 out of 13 unseen datasets, which can surpass models trained on the other 6 datasets individually by a large gain. We hypothesize that the general forms of table-based QA tasks can encourage models to gain general QA ability, which is beneficial when transferring to other tasks or datasets, since instruction tuning requires models to answer the question in essence. However, the models that are individually trained on other tasks may have learned strong superficial regularities as their formats have unique characteristics specially designed for themselves. Therefore, when evaluating on other unseen datasets or tasks, the models are too obfuscated to generate the correct answer.

\textit{Incorporating other tasks helps enhance the model's underlying generalization ability within the same task category.} Comparing the model trained on TabFact and \texttt{TableInstruct}, when evaluating on FEVEROUS, which is the same task transfer for TabFact, we found \texttt{TableLlama} achieves 73.77 accuracy while the model trained on TabFact only achieves 56.15 accuracy. This indicates that other tasks in the training set also play an important role in engaging the model to obtain stronger table fact verification ability. Besides, if we compare the performance on three out-of-domain table QA datasets (i.e., HybridQA, WikiSQL and WikiTQ) among \texttt{TableLlama} and models individually trained on two table-based QA datasets (i.e., HiTab and FeTaQA), we can see \texttt{TableLlama} achieves better zero-shot performance. This indicates that including the other tasks (i.e., \texttt{TableInstruct}) to train the model can further enhance the model's underlying table question answering ability.

 \textit{Individually fine-tuning models on tasks that are highly different from others tends to make models overfit and hardly generalize to others.}\nop{\hs{maybe this is the place to talk about `task family'? do we observer better transfer performance in the same task family? seems no?}} As Table \ref{tab:entire_ablation} shows, the model individually fine-tuned on 4 tasks: column type annotation, relation extraction, entity linking and TabFact tends to have weaker performance when evaluated on other tasks. We hypothesize that these four tasks are highly different from others, so the model individually trained on such tasks will overfit to the task itself, thus becoming hard to generalize to other unseen tasks.


\nop{\hs{you need to comment on the gap from SoTA and cannot avoid that. You may point out that you are zero-shot and theirs is full-dataset training. encourage future work to further improve.}}
\section{Related Work}

\noindent\textbf{Table Representation Learning.}  
Given the vast amount of knowledge stored in tables, various table-based tasks have been proposed \cite{pujara2021tables}, such as column type annotation \citep{hulsebos2019sherlock}, row population \citep{zhang2017entitables}, table QA \cite{sun2016table, wikitq, cheng-etal-2022-hitab, Nan2021FeTaQAFT}, etc. In order to handle the semi-structured tables, existing work puts their efforts into designing special model architectures, \nop{\hs{for the work you are going to mention, if they have a model name, mention it, as it could make people easily recall the work.}} such as TURL with structure-aware attention \citep{deng2020turl}, TUTA with tree-based attention \citep{wang2021tuta} and TaBERT with vertical self-attention mechanism \citep{yin-etal-2020-tabert}; or designing special encodings such as \nop{cell text encoding \citep{yin-etal-2020-tabert, eisenschlos2021mate, wang2021tuta}, }table position encoding \citep{herzig-etal-2020-tapas, wang2021tuta}, and numerical encoding \citep{wang2021tuta} to better encode the table structure and infuse more information to the neural architecture. In addition, some work focuses on table pretraining \citep{liu2022tapex, yin-etal-2020-tabert, deng2020turl, iida-etal-2021-tabbie} to encode knowledge in large-scale tables. However, although such existing works have shown promising progress, they are still data-specific and downstream task-specific, which requires special design tailored for tables and table-based tasks. 

Our work proposes \texttt{TableInstruct} to unify different table-based tasks and develops a one-for-all LLM \texttt{TableLlama} to reduce those extra efforts during modeling\nop{, and evaluate its table understanding and generalization ability under both in-domain and out-of-domain settings}.
This high-level insight is similar to UnifiedSKG \citep{xie2022unifiedskg}, which unifies a diverse set of structured knowledge grounding tasks into a text-to-text format. \nop{and enhance T5 model's performance via multi-task fine-tuning. }However, UnifiedSKG deals with different knowledge sources such as databases, knowledge graphs and web tables and does not explore instruction tuning, while we focus on a wide range of realistic tasks based on real-world tables via instruction tuning.
In addition, a concurrent work \citep{li2023tablegpt} synthesizes diverse table-related tasks and finetunes close-source LLMs such as GPT-3.5 via instruction tuning. \nop{\hs{rephrase? what do you mean? unify diverse table-based tasks on closed-source LLMs such as ChatGPT,}} Compared to theirs, we collect more realistic and complex task data such as HiTab as well as classification and ranking tasks with candidates from Freebase and Wikidata and develop open-source LLMs for table-based tasks. We believe both our constructed high-quality table instruction tuning dataset and the trained model can be valuable resources for facilitating this line of research.

\noindent\textbf{Instruction Tuning.}
Instruction tuning that trains LLMs using $<$instruction, output$>$ pairs in a supervised fashion is a crucial technique to enhance the capabilities and controllability of LLMs \citep{chung2022scaling, wang-etal-2022-super, mishra-etal-2022-cross}. The instructions serve to constrain the model's outputs to align with the desired response characteristics or domain knowledge and can help LLMs rapidly adapt to a specific domain without extensive retraining or architecture designs \citep{zhang2023instruction}. Therefore, different instruction tuning datasets have been proposed to guide LLMs' behaviors \citep{wang-etal-2022-super, honovich2022unnatural, longpre2023flan, xu2023wizardlm, yue2024mammoth}. \nop{Those datasets are collected either from formatting existing natural language processing tasks by templates \citep{longpre2023flan} or prompting ChatGPT \citep{xu2023wizardlm} and GPT-4 \citep{gpt4llm} to generate instructions.}
Different instruction tuning models such as InstructGPT \citep{instructgpt}, Vicuna \citep{vicuna} and Claude\footnote{https://www.anthropic.com/index/introducing-claude}\nop{\hs{can you use another way to cite Claude? it's weird to have the first footnote towards the end of the paper. If you cannot find a paper, just create a bibtex for this url.}} emerge and demonstrate boosted performance compared with the pre-trained models. In addition, instruction tuning has been applied to different modalities such as images, videos and audio \citep{li2023blip2} and has shown promising results. This signals that instruction tuning can be a promising technique to enable large pre-trained models to handle various tasks. However, how to utilize instruction tuning to guide LLMs to complete tables-based tasks is still under-explored. Our work fills this gap by constructing a high-quality table instruction tuning dataset: \texttt{TableInstruct}, which covers large-scale diverse and realistic tables and tasks to enable both modeling and evaluation. We also release \texttt{TableLlama}, an open-source LLM-based generalist model fine-tuned on \texttt{TableInstruct} to promote this avenue of research.
\section{Conclusion}

This paper makes the first step towards developing open-source large generalist models for a diversity of table-based tasks. Towards that end, we construct
\texttt{TableInstruct} and develop the first open-source generalist model for tables, \texttt{TableLlama}.\nop{, a comprehensive dataset
for instructing tuning and evaluating LLMs for
tables and develop the first open-source generalist model for tables, \texttt{TableLlama}, by fine-tuning Llama 2 (7B)
with LongLoRA to address the context length challenge.} We evaluate
both in-domain and out-of-domain settings and the experiments show that \texttt{TableLlama} has gained strong table understanding ability and generalization ability. \nop{On 7 out of 8 in-domain tasks, our generalist model \texttt{TableLlama} achieves comparable or better performance than the existing SOTA method for each task, despite the latter often has table-specific model
design or pre-training. On 6 out-of-domain datasets, it
achieves 6-48 absolute point gains compared
with the base model, showing that training on
our \texttt{TableInstruct} enhances generalizability.}

\section{Limitations}

Although we strive to increase the diversity of our dataset and have collected 14 datasets of 11 tasks for tables, there are still some table-based tasks such as data imputation and table classification which are not included in \texttt{TableInstruct}. Therefore, even if \texttt{TableLlama} has demonstrated the generalization ability on different out-of-domain datasets and tasks, the model's performance may vary based on the complexity and specifics of the new unseen table tasks and datasets. As we have made the first step towards building an open large generalist model for tables, we encourage future work to further explore this line of research and to further enhance the model's generalization ability for tables.
\section*{Acknowledgements}
The authors would thank all members of the OSU NLP group for providing feedback about the project. This research was sponsored in part by NSF IIS-1815674, NSF CAREER \#1942980, and NSF OAC-2112606. The views and conclusions contained herein are those of the authors and should not be interpreted as representing the official policies, either expressed or implied, of the U.S. government. The U.S. Government is authorized to reproduce and distribute reprints for Government purposes notwithstanding any copyright notice herein.

\bibliography{reference}

\appendix

\newpage

\section{Existing SOTA Models}\label{sota_model}
TURL \citep{deng2020turl} is an encoder-based BERT-like model pre-trained on 570K tables. Though TURL has shown SOTA performance on various table tasks such as column type annotation, relation extraction, entity linking, row population and schema augmentation, it requires fine-tuning task-specific modules on labeled data.\nop{when generalizing to new tasks or new datasets} The SOTA method for HiTab builds on 1) TUTA \citep{wang2021tuta}, which uses tree attention as the encoder to capture table structures and 2) FORTAP \citep{cheng2021fortap}, which leverages
spreadsheet formulas for table pre-training to better handle numerical reasoning. The SOTA method for TabFact designs a self-labeled keypoint alignment \citep{zhao-yang-2022-table} to align salient evidence and aggregate essential information between the statement and table. For HybridQA, the SOTA method MAFiD \citep{lee-etal-2023-mafid} deploys special fusion in decoder and uses a gated cross-attention
layer to enhance the reasoning ability on tables. The SOTA method for WikiTQ is TAPEX \citep{liu2022tapex}, which fuses table pre-training by learning a neural SQL executor over a synthetic corpus. The SOTA method for WikiSQL uses two denoising objectives and a clause-sensitive execution guided (EG) decoding strategy to generate better SQL and then get the answer \citep{xu2023sead}. For FeTaQA, KVRET and ToTTo, the SOTA results come from T5-3B fine-tuned on their own individual training data \citep{xie2022unifiedskg}. For FEVEROUS, the SOTA is from a 20B large language model: FLAN UL2 \cite{tay2022ul2}.
\nop{\hs{is this from unifiedSKG? why there is no citation?}. }

\section{More details about \texttt{TableInstruct}}

\subsection{Data Selection}

We choose the datasets and tasks based on three criteria: diversity, realisticness and reliability.
\begin{itemize}
    \item Diversity: we hope to cover table-based tasks as comprehensively as possible both in the NLP community and database community. That’s why we include 14 datasets of 11 tasks.
    \item Realisticness: we include the table sources from Wikipedia tables and National Science Foundation reports (eg, https://www.nsf.gov/statistics/2019/nsf19319/), which make sure the table types are realistic and include both simple tables and hierarchical tables with complex table structures.
    \item Reliability: we compile existing datasets that are widely used in the NLP community and database community.
\end{itemize}
We split TableInstruct into in-domain (for training and evaluation) and out-of-domain (for evaluation) sets based on three constraints:
\begin{itemize}
\item to make the tasks in the training and out-of-domain evaluation set as disjoint as possible;
\item if there are two datasets for the same task, we will divide them into training set and out-of-domain evaluation set;
\item since tables have special two-dimensional structures, we need the model to gain fundamental table understanding abilities, which the model can recognize the relation for cells within and among different columns and rows, and also correlate the headers and row names with corresponding columns and rows. So we mainly select different table interpretation and table augmentation tasks to encourage the model to understand table structures. In addition, we try to engage the model with strong numerical reasoning ability, open-ended table QA and fact verification ability, so we include HiTab, FeTaQA and TabFact for training as well. For out-of-domain tasks, we mainly test the more high-level ability to see the model's generalization. For example, the table question answering datasets in the training set are two types: one is full of numerical reasoning on hierarchical tables and the other is to generate open-ended answer based on highlighted table cells. We hope the learned table QA ability can transfer to different kinds of unseen table QA tasks such as adding extra components (passages or dialogues, etc) as evidence and letting the model infer the answer from both tables and added components.
\end{itemize}

\subsection{Data Annotation}
The raw tables in our collected datasets are stored in JSON, CSV or text files. We mainly annotate instructions and questions based on the metadata of each task, serialize the table format and put the ground truth as response (more details and example cases are in Appendix \ref{sec:prompt_format}).

\subsection{Quality Control}
These collected datasets are cleaned by previous authors. After we annotated the data, we randomly sampled 30 instances for each task to double check the data and make sure there are no errors. We also have two annotators to do the cross-checking.

\section{More detailed statistics of \texttt{TableInstruct}.}
Table \ref{tab:statistic_appendix} shows more detailed statistics of \texttt{TableInstruct} in terms of the average word count of different parts of the
datasets (i.e., instruction, input, question and response), table size (average column size and row size per table), table
type (Wikipedia tables or NSF reports), task type (ranking or classification) and whether the tables are hierarchical
or not.

\onecolumn

\begin{table*}[]
\small
\centering
\resizebox{\linewidth}{!}{
\begin{tabular}{@{}lccccccccccccc@{}}
\toprule
& & & Avg & & Avg & Avg & & & & Hierarchical & Hierarchical\\
& Avg & Avg & Instruction & Avg Input & Question & Response & Table & & & Col & Row \\
& Rows/Table & Cols/Table & Len(Word) & Len(Word) & Len(Word) & Len(Word) & Type & Ranking? & Classification? & Headers? & Headers? \\
\midrule
\multicolumn{12}{c}{In-domain} \\
\midrule
ColType & 15 & 7 & 46 & 374 & 333 & 2 & Wiki. & N & Y & N & N \\
RelExtra & 18 & 7 & 45 & 433 & 245 & 1 & Wiki. & N & Y & N & N \\
EntLink & 60 & 6 & 82 & 1308 & 2070 & 9 & Wiki. & N & Y & N & N \\
ScheAug & - & - & 51 & 17 & 24 & 12 & Wiki. & Y & N & N & N \\
RowPop & - & - & 60 & 52 & 74 & 62 & Wiki. & Y & N & N & N \\
HiTab & 22 & 9 & 29 & 491 & 17 & 1 & Stat. reports \& Wiki. & N & N & Y & Y \\
FeTaQA & 15 & 6 & 28 & 325 & 39 & 19 & Wiki. & N & N & Y & Y \\
TabFact & 14 & 6 & 27 & 315 & 27 & 1 & Wiki. & N & Y & N & N \\
\midrule
\multicolumn{12}{c}{Out-of-domain} \\
\midrule
FEVER. & 13 & 4 & 27 & 362 & 63 & 1 & Wiki. & N & Y & Y & Y \\
HybridQA & 15 & 4 & 21 & 315 & 19 & 2 & Wiki. & N & N & N & N \\
KVRET & 7 & 6 & 55 & 171 & 46 & 9 & Wiki. & N & N & N & N \\
ToTTo & 32 & 7 & 21 & 54 & 13 & 15 & Wiki. & N & N & Y & Y \\
WikiSQL & 15 & 6 & 19 & 285 & 12 & 2 & Wiki. & N & N & N & N \\
WikiTQ & 19 & 6 & 19 & 348 & 10 & 2 & Wiki. & N & N & N & N \\
\bottomrule
\end{tabular}
}
\caption{More detailed statistics of \texttt{TableInstruct} in terms of the average word count of different parts of the datasets (i.e., instruction, input, question and response), table size (average column size and row size per table), table type (Wikipedia tables or NSF reports), task type (ranking or classification) and whether the tables are hierarchical or not. 'Y' indicates 'Yes' and 'N' indicates 'No'.}
\label{tab:statistic_appendix}
\end{table*}

\section{Case Study}
\label{sec:case_study}
\begin{table}[htbp]
\centering
\resizebox{\linewidth}{!}{
\begin{tabular}{lcccccc}
\toprule
Query Caption&Seed&Candidates&Target&AP&Predicted\\\midrule
concord quarry dogs &2002\_NECBL\_season & \begin{tabular}[c]{@{}c@{}}2003\_Amsterdam\_Admirals\_season\\
The\_Young\_Punx\\2011\_FCBL\_season\\...\end{tabular}&{
\begin{tabular}[c]{@{}c@{}}2003\_NECBL\_season\\ 2004\_NECBL\_season\\
2005\_NECBL\_season\\
2006\_NECBL\_season\end{tabular}
} &1.0 & {
\begin{tabular}[c]{@{}c@{}}2003\_NECBL\_season\\ 2004\_NECBL\_season\\
2005\_NECBL\_season\\
2006\_NECBL\_season\end{tabular}
} \\ 
 \midrule
 {\begin{tabular}[c]{@{}c@{}}oleg veretelnikov\\achievements\end{tabular} }& \begin{tabular}[c]{@{}c@{}}1993\_Asian\_Athletics\\\_Championships\\
 \end{tabular}
 &{\begin{tabular}[c]{@{}c@{}}New\_York\_City\_Marathon\\
 Friendship\_Games\\
1998\_Asian\_Games\\ 
...\\
\end{tabular}
} & {
\begin{tabular}[c]{@{}c@{}} {\begin{tabular}[c]{@{}c@{}}1997\_World\_Championships\_in\\\_Athletics-2013\_Men's\_decathlon\\ \end{tabular}
}\\
1994\_Asian\_Games\\
1999\_World\_Championships\_in\_Athletics\\
1998\_Asian\_Games\\
\end{tabular}
}  & 0.2 &{\begin{tabular}[c]{@{}c@{}}1994\_Asian\_Games\\1995\_Asian\_Athletics\_Championships\\ Athletics\_at\_the\_1995\_All-Africa\_Games\\
...
\end{tabular}
}\\ 
\bottomrule 
\end{tabular}
}
\caption{Case study for row population task. ``Query Caption" refers to the table metadata such as Wikipedia page title and table caption. ``AP" means average precision.}
\label{tab:row_pop}
\end{table}

\section{Example Prompts}

\label{sec:prompt_format}


\begin{figure*}[htbp]
\begin{tcolorbox}[colback=green!7.5!white, colframe=green!40!black, title=Column Type Annotation, fontupper=\footnotesize, fonttitle=\footnotesize]
Below is an instruction that describes a task, paired with an input that provides further context. Write a response that appropriately completes the request. \\

\textbf{\#\#\# Instruction}: \\
This is a column type annotation task. The goal for this task is to choose the correct types for one selected column of the table from the given candidates. The Wikipedia page, section and table caption (if any) provide important information for choosing the correct column types. \\

\textbf{\#\#\# Input}: \\
\text{[TLE]} The Wikipedia page is about 1958 Nippon Professional Baseball season. The Wikipedia section is about Central League. The table caption is Pitching leaders. [TAB] col: | stat | player | team | total | [SEP] row 1: | Wins | Masaichi Kaneda | Kokutetsu Swallows | 31| [SEP] row 2: | Losses | Noboru Akiyama | ... \\

\textbf{\#\#\# Question}: \\
The column 'player' contains the following entities: <Masaichi Kaneda>, <Noboru Akiyama>, etc. \textcolor{red}{The column type candidates are: tv.tv\_producer, astronomy.star\_system\_body, location.citytown, sports.pro\_athlete, biology.organism, medicine.muscle, baseball.baseball\_team, baseball.baseball\_player, aviation.aircraft\_owner, people.person, ...} What are the correct column types for this column (column name: player; entities: <Masaichi Kaneda>, <Noboru Akiyama>, etc)? \\

\textbf{\#\#\# Response}: \\
sports.pro\_athlete, baseball.baseball\_player, people.person.

\end{tcolorbox}
\caption{\textbf{Column type annotation} task. This task is to annotate the selected column with the correct semantic types. We mark candidates with \textcolor{red}{red color} in the "task instruction" part. The candidate size can be up to hundreds to thousands in \texttt{TableInstruct}.}
\label{fig:col_type}
\end{figure*}


\begin{figure*}[htbp]
\begin{tcolorbox}[colback=green!7.5!white, colframe=green!40!black, title=Relation Extraction, fontupper=\footnotesize, fonttitle=\footnotesize]
Below is an instruction that describes a task, paired with an input that provides further context. Write a response that appropriately completes the request. \\

\textbf{\#\#\# Instruction}: \\
This is a relation extraction task. The goal for this task is to choose the correct relations between two selected columns of the table from the given candidates. The Wikipedia page, section and table caption (if any) provide important information for choosing the correct relation types. \\

\textbf{\#\#\# Input}: \\
\text{[TLE]} The Wikipedia page is about Yukon Legislative Assembly. The Wikipedia section is about Current members. [TAB] col: | | name | party | riding | row 1: | | Kevin Barr | New Democratic Party | Mount Lorne-Southern Lakes | [SEP] row 2: | | Brad Cathers | ... \\

\textbf{\#\#\# Question}: \\
The two selected column names are: <(name),(party)>. The entity pairs for these two columns are: <(Kevin Barr),(New Democratic Party)>, <(Brad Cathers),(Yukon Party)>, <(Currie Dixon),(Yukon
Party)>, <(Darius Elias),(Yukon Party)>, ... \textcolor{red}{The relation type candidates are: location.location.contains, aviation.airline.hubs, film.film.written\_by, time.event.instance\_of\_recurring\_even , people.person.place\_of\_birth, music.composer.compositions, sports.sports\_team.roster- sports.sports\_team\_roster.player, location.location.containedby, soccer.football\_player.statistics- soccer.football\_player\_stats.team...} What are the correct relation types for the two selected columns (column names: <(name),(party)>. entity pairs:  <(Kevin Barr),(New Democratic Party)>, <(Brad Cathers),(Yukon Party)>, <(Currie Dixon), (Yukon Party)>, <(Darius Elias),(Yukon Party)>, etc)? \\

\textbf{\#\#\# Response}: \\
government.politician.party-government.political\_party\_tenure.party.

\end{tcolorbox}
\caption{\textbf{Relation extraction} task. This task is to annotate the selected column pairs with the correct relations. We mark candidates with \textcolor{red}{red color} in the "task instruction" part. The candidate size can be up to hundreds to thousands in \texttt{TableInstruct}.}
\label{fig:rel_extraction}
\end{figure*}


\begin{figure*}[htbp]
\begin{tcolorbox}[colback=green!7.5!white, colframe=green!40!black, title=Entity Linking, fontupper=\footnotesize, fonttitle=\footnotesize]
Below is an instruction that describes a task, paired with an input that provides further context. Write a response that appropriately completes the request. \\

\textbf{\#\#\# Instruction}: This is an entity linking task. The goal for this task is to link the selected entity mention in the table cells to the entity in the knowledge base. You will be given a list of referent entities, with each one composed of an entity name, its description and its type. Please choose the correct one from the referent entity
candidates. Note that the Wikipedia page, Wikipedia section and table caption (if any) provide important
information for choosing the correct referent entity. \\

\textbf{\#\#\# Input}: [TLE] The Wikipedia page is about A-League all-time records. The Wikipedia section is about Average season attendances. [TAB] col: | season | league average | total gate receipts | highest club | average |
lowest club | average | row 1: | 2005-06 | 10,955 | 920,219 | Sydney FC | 16,669 | New Zealand Knights | 3,909 | [SEP] row 2: | 2006-07 | 12,927 | ... \\

\textbf{\#\#\# Question}: The selected entity mention in the table cell is: Melbourne Victory. The column name for 'Melbourne Victory' is highest club. \textcolor{red}{The referent entity candidates are: <Melbourne Victory FC W-League [DESCRIPTION] None [TYPE] SoccerClub>, <2016-17 Melbourne Victory FC season [DESCRIPTION] None [TYPE] SoccerClubSeason>, <2011-12 Melbourne Victory season [DESCRIPTION] Association football club 2011/12 season for Melbourne Victory [TYPE] SoccerClubSeason>, ...} What is the correct referent entity for the entity mention 'Melbourne Victory' ? \\

\textbf{\#\#\# Response}:
<Melbourne Victory [DESCRIPTION] association football team from Australia [TYPE] SoccerClub>.

\end{tcolorbox}
\caption{\textbf{Entity linking} task. This task is to link the selected entity mention in the table cells to the entity in the knowledge base. We mark candidates with \textcolor{red}{red color} in the "task instruction" part. The candidate size can be up to
hundreds to thousands in \texttt{TableInstruct}.}
\label{fig:ent_link}
\end{figure*}


\begin{figure*}[htbp]
\begin{tcolorbox}[colback=green!7.5!white, colframe=green!40!black, title=Row Population, fontupper=\footnotesize, fonttitle=\footnotesize]
Below is an instruction that describes a task, paired with an input that provides further context. Write a response that appropriately completes the request. \\

\textbf{\#\#\# Instruction}: This is a table row population task. The goal of this task is to populate the possible entities of the selected column for a table, given the Wikipedia page title, Wikipedia section title, table caption (if any) and table headers. You will be given a list of entity candidates. Please rank them so that the most likely entities come first. \\

\textbf{\#\#\# Input}: [TLE] The Wikipedia page is about NBA conference finals. The Wikipedia section is about eastern conference finals. The table headers are: | year | champion | coach | result | runner-up | coach |. You need to populate the column: year. [SEED] The seed entity is <1971\_NBA\_playoffs>. \\

\textbf{\#\#\# Question}: \textcolor{red}{The entity candidates are: <2003\_NBA\_playoffs>, <1982-83\_Washington\_Bullets\_season>,
<2004\_NBA\_playoffs>, <Philadelphia\_76ers>, <1983-84\_Washington\_Bullets\_season>,
<1952\_NBA\_playoffs>, <1972\_NBA\_playoffs>, <1999-2000\_Dallas\_Mavericks\_season>, <1985-
86\_Sacramento\_Kings\_season>, <2000-01\_Vancouver\_Grizzlies\_season>, <Toronto\_Raptors>,
<Vancouver\_Grizzlies>, <1976\_NBA\_playoffs>, ...} \\

\textbf{\#\#\# Response}:
<1972\_NBA\_playoffs>, <1973\_NBA\_playoffs>, <1974\_NBA\_playoffs>, <1975\_NBA\_playoffs>,
<1976\_NBA\_playoffs>, <1977\_NBA\_playoffs>, ...

\end{tcolorbox}
\caption{\textbf{Row population} task. This task is to populate the possible entities of the selected column for a table given partial table and table metadata. We mark candidates with \textcolor{red}{red color} in the  "task instruction" part. The candidate size can be up to hundreds to thousands in \texttt{TableInstruct}.}
\label{fig:row_pop}
\end{figure*}


\begin{figure*}[htbp]
\begin{tcolorbox}[colback=green!7.5!white, colframe=green!40!black, title=Schema Augmentation, fontupper=\footnotesize, fonttitle=\footnotesize]
Below is an instruction that describes a task, paired with an input that provides further context. Write a response that appropriately completes the request. \\

\textbf{\#\#\# Instruction}: \\
This is a table schema augmentation task. The goal of this task is to populate the possible headers for a table, given the table caption and the seed table header. You will be given a list of table header
candidates. Please rank them so that the most likely headers come first. \\

\textbf{\#\#\# Input}: \\
\text{[TLE]} The table caption is 2010-11 rangers f.c. season. [SEED] The seed table header is <competition>. \\

\textbf{\#\#\# Question}: \\
\textcolor{red}{The header candidates are: <from>, <fee (\textbackslash{u}00a3)>, <opponents>, <final position / round>, <started
round>, <player>, <fee>, <scorers>, <position (s)>, <name>, <venue>. Please rank the headers in the header candidates.} \\

\textbf{\#\#\# Response}: \\
<town/city>, <summary>, <suburb>, <county>, <region>, <district>, <stadium>, <city>, <home team>,
<remarks>, <city name>, <film name> \\


\end{tcolorbox}
\caption{\textbf{Schema augmentation} task. This task is to populate the possible headers for a table, given the table caption and the seed table header. The targets in this case are: <started round>, <final position/round>.}
\label{fig:schema_aug}
\end{figure*}


\begin{figure*}[htbp]
\begin{tcolorbox}[colback=green!7.5!white, colframe=green!40!black, title=Hierarchical Table QA, fontupper=\footnotesize, fonttitle=\footnotesize]
Below is an instruction that describes a task, paired with an input that provides further context. Write a response that appropriately completes the request. \\

\textbf{\#\#\# Instruction}: This is a hierarchical table question answering task. The goal for this task is to answer the given question based on the given table. The table might be hierarchical. \\

\textbf{\#\#\# Input}: [TLE] The table caption is department of defense obligations for research, development, test, and evaluation, by agency: 2015-18. [TAB] | agency | 2015 | 2016 | 2017 | 2018 | [SEP] | department of defense | department of defense | department of defense | department of defense | department of defense | [SEP] | rdt\&e | 61513.5 | 69306.1| 70866.1 | 83725 | [SEP] | total research | 6691.5 | 7152 | 7178 | 7652.7 | [SEP] | basic research | 2133.4 | 2238.7 | 2110.1 | 2389.9 | [SEP] | defense advanced research projects agency | defense advanced research projects agency | defense advanced research projects agency | ... \\

\textbf{\#\#\# Question}: How many dollars are the difference for total research of department of the air force increase between 2016 and 2018? \\

\textbf{\#\#\# Response}: \\
142.3. \\

\end{tcolorbox}
\caption{\textbf{Hierarchical table QA} task. This task is to answer the question based on the tables with complex hierarchical structures.}
\label{fig:hitab}
\end{figure*}


\begin{figure*}[htbp]
\begin{tcolorbox}[colback=green!7.5!white, colframe=green!40!black, title=Highlighted Cells QA, fontupper=\footnotesize, fonttitle=\footnotesize]
Below is an instruction that describes a task, paired with an input that provides further context. Write a response that appropriately completes the request. \\

\textbf{\#\#\# Instruction}: \\
This is a free-form table question answering task. The goal for this task is to answer the given question
based on the given table and the highlighted cells. \\

\textbf{\#\#\# Input}: \\
\text{[TLE]} The Wikipedia page title of this table is Holly Dunn. The Wikipedia section title of this table is Singles. [TAB] | Year | Single | Peak chart positions | Peak chart positions | Album | [SEP] | Year | Single | US Country | CAN Country | Album | [SEP] | 1985 | ... \\

\textbf{\#\#\# Question}: \\
The highlighted cells of the table are: [HIGHLIGHTED\_BEGIN] [1988], [Across the Rio Grande in 1988 included the singles \textbackslash{"}That's What Your Love Does to Me\textbackslash{"} and \textbackslash{"}(It's Always Gonna Be) Someday\textbackslash{"}.], [\textbackslash{"}That's What Your Love Does to Me\textbackslash{"}], [Across the Rio Grande], [1988], [\textbackslash{"}(It's Always Gonna Be) Someday\textbackslash{"}], [Across the Rio Grande] \text{[HIGHLIGHTED\_END]} What singles were Included in Across the Rio Grande in 1988? \\

\textbf{\#\#\# Response}: \\
Across the Rio Grande in 1988 included the singles \textbackslash{"}That's What Your Love Does to Me\" and \textbackslash{"}(It's Always Gonna Be) Someday\textbackslash{"}. \\

\end{tcolorbox}
\caption{\textbf{Highlighted cells QA} task. This task is to answer the question based on the tables with highlighted cells.}
\label{fig:fetaqa}
\end{figure*}


\begin{figure*}[htbp]
\begin{tcolorbox}[colback=green!7.5!white, colframe=green!40!black, title=Table Fact Verification (TabFact), fontupper=\footnotesize, fonttitle=\footnotesize]
Below is an instruction that describes a task, paired with an input that provides further context. Write a response that appropriately completes the request. \\

\textbf{\#\#\# Instruction}: \\
This is a table fact verification task. The goal of this task is to distinguish whether the given statement is entailed or refuted by the given table. \\

\textbf{\#\#\# Input}: \\
\text{[TLE]} The table caption is about tony lema. [TAB] | tournament | wins | top - 5 | top - 10 | top - 25 | events |
cuts made [SEP] | masters tournament | 0 | 1 | 2 | 4 | 4 | 4 | [SEP] | us open | 0 | 2 | 3 | 4 | 6 | 5 | [SEP] | the
open championship | 1 | 2 | 2 | 2 | 3 | 3 | [SEP] | pga championship | 0 | 0 | 1 | 2 | 5 | 4 | [SEP] | totals | 1 | 5 | 8
| 12 | 18 | 16 |. \\

\textbf{\#\#\# Question}: \\
The statement is: <tony lema be in the top 5 for the master tournament, the us open, and the open championship>. Is it entailed or refuted by the table above? \\

\textbf{\#\#\# Response}: \\
Entailed. \\

\end{tcolorbox}
\caption{\textbf{Table fact verification} task. This task is to discriminate whether the claim can be entailed or refuted by the given table.}
\label{fig:tabfact}
\end{figure*}


\begin{figure*}[htbp]
\begin{tcolorbox}[colback=green!7.5!white, colframe=green!40!black, title=Hybrid Question Answering, fontupper=\footnotesize, fonttitle=\footnotesize]
Below is an instruction that describes a task, paired with an input that provides further context. Write a response that appropriately completes the request. \\

\textbf{\#\#\# Instruction}: \\
This is a hybrid question answering task. The goal of this task is to
answer the question given tables and passages. \\

\textbf{\#\#\# Input}: \\
\text{[TAB]} col: | rank | player | team (s) by season | carries | yards | average | [SEP] | 1 | emmitt smith | dallas cowboys ( 1990 - 2002 ) arizona cardinals ( | 4,409 | 18,355 | 4.2 | [SEP] | 3 | frank gore | san francisco 49ers ( 2005 - 2014 ) indianapolis colts | 3,548 | 15,347 | 4.3 | [SEP] | ... \\

\textbf{\#\#\# Question}: \\
The passage may also provide related context. You can refer to both the passages and the table when you answer the question. \textbf{Passages}: emmitt smith: smith led the league in rushing and won the super bowl in the same year three times ( 1992 , 1993 , and 1995 ) when to that point it had never been done . | walter payton: walter jerry payton ( july 25 , 1954 - november 1 , 1999 ) was an american professional football player who was a running back for the chicago bears of the national football league ( nfl ) for thirteen seasons . | ... \textbf{The question}: what is the middle name of the player with the second most national football league career rushing yards? \\

\textbf{\#\#\# Response}: \\
Jerry. \\

\end{tcolorbox}
\caption{\textbf{HybridQA} task. This task is to answer the question based on the table and passages.}
\label{fig:hybridqa}
\end{figure*}


\begin{figure*}[htbp]
\begin{tcolorbox}[colback=green!7.5!white, colframe=green!40!black, title=Table Grounded Dialogue Generation, fontupper=\footnotesize, fonttitle=\footnotesize]
Below is an instruction that describes a task, paired with an input that provides further context. Write a response that appropriately completes the request. \\

\textbf{\#\#\# Instruction}: \\
This is a dialogue response generation task grounded on tables. The goal of this task is to generate response based on the given dialogue history and the given table. The dialogues are grounded through underlying tables and span three distinct tasks in the in-car personal assistant space: calendar scheduling, weather information retrieval, and point-of-interest navigation. \\

\textbf{\#\#\# Input}: \\
col : poi | distance | traffic\_info | poi\_type | address row 1 : chevron | 5\_miles | moderate\_traffic | gas\_station | 783\_arcadia\_pl row 2 : town\_and\_country | 5\_miles | no\_traffic | shopping\_center | 383\_university\_ave \\

\textbf{\#\#\# Question}: \\
The dialogue history is: <what is the address ? || taking you to chevron | that s good ! please pick the quickest route to get there and avoid all heavy\_traffic ! | there is a chevron | what gas\_station are here ?>. Please generate the response based on the given table and the given dialogue history. \\

\textbf{\#\#\# Response}: \\
783\_arcadia\_pl is the address for chevron gas\_station. \\

\end{tcolorbox}
\caption{\textbf{Table grounded dialogue generation} task. This task is to generate the response based on the given table and the dialogue history.}
\label{fig:table_grounded_dialogue_generation}
\end{figure*}


\begin{figure*}[htbp]
\begin{tcolorbox}[colback=green!7.5!white, colframe=green!40!black, title=Highlighted Cells Description, fontupper=\footnotesize, fonttitle=\footnotesize]
Below is an instruction that describes a task, paired with an input that provides further context. Write a response that appropriately completes the request. \\

\textbf{\#\#\# Instruction}: \\
This is a highlighted cells description task. The goal of this task is to generate the language description given table cells. \\

\textbf{\#\#\# Input}: \\
<page\_title> List of Governors of South Carolina </page\_title> <section\_title> Governors under the Constitution of 1868 </section\_title> <table> <cell> 76 <col\_header> \# </col\_header> <col\_header> 74 </col\_header> <col\_header> 75 </col\_header> </cell> <cell> Daniel Henry Chamberlain <col\_header> Governor </col\_header> <row\_header> 76 </row\_header> </cell> <cell> December 1, 1874 <col\_header>
Took Office </col\_header> <row\_header> 76 </row\_header> </cell> </table>. \\

\textbf{\#\#\# Question}: \\
Please generate one natural language description to describe the given highlighted table cells. \\

\textbf{\#\#\# Response}: \\
Daniel Henry Chamberlain was the 76th Governor of South Carolina from 1874. \\

\end{tcolorbox}
\caption{\textbf{Highlighted cells description} task. This task is to generate the language description for the highlighted table cells.}
\label{fig:totto}
\end{figure*}


\begin{figure*}[htbp]
\begin{tcolorbox}[colback=green!7.5!white, colframe=green!40!black, title=Table Fact Verification (FEVEROUS), fontupper=\footnotesize, fonttitle=\footnotesize]
Below is an instruction that describes a task, paired with an input that provides further context. Write a response that appropriately completes the request. \\

\textbf{\#\#\# Instruction}: \\
This is a table fact verification task. The goal of this task is to distinguish whether the given statement is entailed or refuted by the given table. \\

\textbf{\#\#\# Input}: \\
\text{[TAB]} col: | logical system | lindenbaum\u2013tarski algebra | [SEP] | classical sentential logic | boolean algebra | [SEP] | intuitionistic propositional logic | heyting algebra | [SEP] | ... \\

\textbf{\#\#\# Question}: \\
The statement is: <algebraic logic has five logical system and lindenbaum\u2013tarski algebra which includes physics algebra and nodal algebra (provide models of propositional modal logics).>. Is it entailed or refuted by the table above? If you think the current information can not provide enough evidence for determining it, please choose 'not enough info', otherwise please choose the answer from 'supports' or 'refutes'.  \\

\textbf{\#\#\# Response}: \\
Refutes. \\

\end{tcolorbox}
\caption{\textbf{Table fact verification} task. This task is to discriminate whether the claim can be entailed or refuted by the given table.}
\label{fig:examplars}
\end{figure*}


\begin{figure*}[htbp]
\begin{tcolorbox}[colback=green!7.5!white, colframe=green!40!black, title=Table QA (WikiSQL), fontupper=\footnotesize, fonttitle=\footnotesize]
Below is an instruction that describes a task, paired with an input that provides further context. Write a response that appropriately completes the request. \\

\textbf{\#\#\# Instruction}: \\
This is a table QA task. The goal of this task is to answer the question given the table. \\

\textbf{\#\#\# Input}: \\
\text{[TAB]} col: | player | no. | nationality | position | years in toronto | school/club team | [SEP] | aleksandar radojevi\u0107 | 25 | serbia | center | 1999-2000 | barton cc (ks) | [SEP] | shawn respert | 31 | united states | guard | 1997-98 | michigan state | [SEP] | ... \\

\textbf{\#\#\# Question}: \\
What is terrence ross' nationality? \\

\textbf{\#\#\# Response}: \\
United states. \\

\end{tcolorbox}
\caption{\textbf{Table QA} task. This task is to answer the question based on the given table.}
\label{fig:wikisql}
\end{figure*}


\begin{figure*}[htbp]
\begin{tcolorbox}[colback=green!7.5!white, colframe=green!40!black, title=Table QA (WikiTQ), fontupper=\footnotesize, fonttitle=\footnotesize]
Below is an instruction that describes a task, paired with an input that provides further context. Write a response that appropriately completes the request. \\

\textbf{\#\#\# Instruction}: \\
This is a table QA task. The goal of this task is to answer the question given the table. \\

\textbf{\#\#\# Input}: \\
\text{[TAB]} col: | series \# | season \# | title | notes | original air date | [SEP] | 1 | 1 | \"the charity\" | alfie, dee dee, and melanie are supposed to be helping | october 15, 1994 | [SEP] | 2 | 1 | \"the practical joke war\" | alfie and goo unleash harsh practical jokes on dee dee | october 22, 1994 | [SEP] | ... \\

\textbf{\#\#\# Question}: \\
Alfie's birthday party aired on january 19. What was the airdate of the next episode? \\

\textbf{\#\#\# Response}: \\
January 26, 1995. \\

\end{tcolorbox}
\caption{\textbf{Table QA} task. This task is to answer the question based on the given table.}
\label{fig:wikitq}
\end{figure*}

\end{document}